\documentclass[runningheads]{llncs} 

\usepackage{float}
\usepackage{graphicx}
\usepackage{wrapfig}
\usepackage{multirow}
\usepackage[labelfont=bf]{caption}
\usepackage{color}
\usepackage[colorlinks=true,linkcolor=blue,citecolor=blue]{hyperref}

\usepackage{color}
\usepackage{amsmath} 
\usepackage{mathtools} 
\usepackage{amsfonts} 
\usepackage{amssymb}
\usepackage[normalem]{ulem}

\usepackage{array} 
\usepackage{booktabs}
\usepackage{rotating}
\usepackage{multirow}
\usepackage{multicol}
\usepackage{lscape}
\usepackage{subfig}
\usepackage{comment}
\usepackage[export]{adjustbox}

\usepackage{algorithm,algorithmicx,algpseudocode}

\usepackage{color}

\usepackage[colorlinks=true,linkcolor=blue,citecolor=blue]{hyperref}

\usepackage{tikz,xcolor,hyperref}

\definecolor{lime}{HTML}{A6CE39}
\DeclareRobustCommand{\orcidicon}{
	\begin{tikzpicture}
	\draw[lime, fill=lime] (0,0) 
	circle [radius=0.16] 
	node[white] {{\fontfamily{qag}\selectfont \tiny ID}};
	\draw[white, fill=white] (-0.0625,0.095) 
	circle [radius=0.007];
	\end{tikzpicture}
	\hspace{-2mm}
}

\foreach \x in {A, ..., Z}{\expandafter\xdef\csname orcid\x\endcsname{\noexpand\href{https://orcid.org/\csname orcidauthor\x\endcsname}
			{\noexpand\orcidicon}}
}
			
\usepackage{color}

\definecolor{darkgreen}{rgb}{0.53, 0.66, 0.42}

\begin{document}

\title{Strongly Topology-preserving GNNs for Brain Graph Super-resolution}

\titlerunning{Strongly Topology-preserving GNNs for Brain Graph Super-resolution}  

\author{Pragya Singh\index{Singh, Pragya} \and Islem Rekik\orcidA{}\index{Rekik, Islem}\thanks{ {corresponding author: i.rekik@imperial.ac.uk, \url{http://basira-lab.com}, GitHub: \url{http://github.com/basiralab}. }}}

\authorrunning{P Singh et al.}

\institute{BASIRA Lab, Imperial-X and Department of Computing, Imperial College London, United Kingdom}

\maketitle              

\begin{abstract}

Brain graph super-resolution is an under-explored yet highly relevant task in network neuroscience. It circumvents the need for costly and time-consuming medical imaging data collection, preparation, and processing. Current super-resolution methods leverage graph neural networks (GNNs) thanks to their ability to natively handle graph-structured datasets. However, most GNNs perform node feature learning, which presents two significant limitations: (1) they require computationally expensive methods to learn complex node features capable of inferring connectivity strength or edge features, which do not scale to larger graphs; and (2) computations in the node space fail to adequately capture higher-order brain topologies such as cliques and hubs. However, numerous studies have shown that brain graph \emph{topology} is crucial in identifying the onset and presence of various neurodegenerative disorders like Alzheimer’s and Parkinson’s. Motivated by these challenges and applications, we propose our \textbf{S}trongly \textbf{T}opology-\textbf{P}reserving GNN framework for Brain \textbf{G}raph \textbf{S}uper-\textbf{R}esolution (\textbf{STP-GSR}). It is the first graph super-resolution architecture to perform representation learning in higher-order topological space. Specifically, using the primal-dual graph formulation from graph theory, we develop an efficient mapping from the edge space of our low-resolution (LR) brain graphs to the node space of a high-resolution (HR) dual graph. This approach ensures that node-level computations on this dual graph correspond naturally to edge-level learning on our HR brain graphs, thereby enforcing strong topological consistency within our framework. Additionally, our framework is GNN layer agnostic and can easily learn from smaller, scalable GNNs, significantly reducing computational requirements. We comprehensively benchmark our framework across seven key topological measures and observe that it significantly outperforms the previous state-of-the-art methods and baselines. The STP-GSR code is available at \url{https://github.com/basiralab/STP-GSR}. 
\end{abstract}

\section{Introduction}

Multi-resolution neuroimaging offers rich and complementary information,  playing a crucial role in various neuroscientific studies \cite{chen2018efficient,marek2022reproducible}. Different resolutions offer detailed insights into the brain's anatomical and functional properties, which are instrumental for early disease diagnosis \cite{senan2022early,yen2023exploring}. However, super-resolution imaging datasets are expensive and time-consuming to acquire and process. To address these challenges, numerous deep learning methods have been developed for cross-resolution image synthesis \cite{de2022deep,zhang2021mr,wu2023super}. For instance, \cite{zhang2021mr} utilizes an attention-based architecture to super-resolve PD, T1-w, and T2-w MRI, while \cite{wu2023super} employs diffusion models \cite{ho2020denoising} for T1-w and DWI. Recently, deep learning methods have also achieved great success in modeling graphs and higher-order relational structures \cite{bronstein2021geometric}, including brain connectomes \cite{bassett2017network,sporns2018graph,liu2023braintgl}. For example, \cite{liu2023braintgl} integrates Graph Convolutional Networks (GCNs) \cite{kipf2016semi} with Long Short-Term Memory (LSTM) networks \cite{hochreiter1997long} for the simultaneous spatio-temporal modeling of brain networks.

Despite the parallel advances in multi-resolution image synthesis and deep graph learning, work on super-resolving brain graphs remains scarce. High-resolution brain graphs are necessary for accurate neuroscientific analysis. For example, \cite{finn2015functional} shows that coarse brain graphs diminish individual variability required for accurate functional connectome fingerprinting. Yet brain graph super-resolution is challenging for several reasons \cite{bessadok2022graph}. First, unlike images, low-resolution (LR) and high-resolution (HR) brain graphs lack a hierarchical structure and cannot use simple local transformations. Second, brain graphs are highly dense, leading to higher computational requirements.

Although challenging, there has been some foundational work in tackling brain graph super-resolution using GNNs \cite{isallari2021brain,pala2021template,mhiri2021non,mhiri2021stairwaygraphnet,ghilea2023replica} and earlier works leveraging machine learning \cite{cengiz2019predicting,mhiri2020brain}. For example, \cite{isallari2021brain} formulates graph super-resolution as a node feature learning task, using a graph U-Net architecture \cite{gao2019graph} to introduce hierarchical structure and a graph Laplacian operator to super-resolve LR graphs. Similar to \cite{isallari2021brain}, \cite{mhiri2021non} follows the node feature learning formulation but employs more expressive NNConv layers \cite{simonovsky2017dynamic} for global graph alignment and then applies a graph-GAN model \cite{wang2018graphgan} to generate HR graphs. Another variant of this work \cite{mhiri2021stairwaygraphnet} presents a multi-resolution GNN model (StairwayGraphNet) for generating brain graphs at increasing scales. \cite{pala2021template} uses representation template graphs in low and high-resolution domains as priors to speed up the training of a super-resolution GNN model. Another work \cite{ghilea2023replica} leveraged federated learning to super-resolve brain graphs when learning from limited data. Even though they achieve impressive results in predicting Dosenbach parcellated rfMRI \cite{dosenbach2010prediction} from T1-w MRI, we note some major limitations. \emph{First}, to learn complex node features, these models use computationally intensive GNN layers and are thus not scalable to large brain graphs. \emph{Second}, by operating predominately in the node space, they offer limited capacity to model higher-order topological structures such as cliques, hubs, etc.

The topological learning limitation forms a serious bottleneck that restricts the utility of such models. Numerous studies \cite{pereira2015aberrant,pereira2016disrupted,khazaee2015identifying,nigro2022role,mijalkov2017braph} have shown that brain graph topology plays a central role in correctly identifying the onset and existence of various neurodegenerative disorders, including the two most frequent ones: Alzheimer’s diseases (AD) and Parkinson’s diseases (PD). Notably, \cite{pereira2016disrupted} observes that different stages of AD show decreasing path length and mean clustering compared to the control group. Similarly, \cite{pereira2015aberrant} analyzes topological measures like clustering coefficient, characteristic path length, and small-worldness from 3T MRI data and concludes that aberrant values are observed for early PD patients. Finally, \cite{nigro2022role} shows a correlation between the loss of hubs in certain brain regions and the appearance of more hubs in others for frontotemporal dementia.

These findings motivate our present research on strongly topology-preserving GNN framework for brain graph super-resolution. Recognizing the inherent limitations of computations in the node space, we propose an approach that uplifts these computations to higher-order edge-space. To achieve this, we adapt the primal-dual graph formulation \cite{gross2018graph} from graph theory, mapping the edges in our brain graph (primal graph) to the nodes of its dual graph. This duality ensures that any node computations in the dual space correspond to edge-level learning on the original graph, allowing us to leverage existing GNNs for edge-based regression. Furthermore, as these dual graphs are highly sparse (more than 97\%), they are computationally efficient and less prone to over-smoothing. Additionally, edges, as higher-order topological objects, more effectively capture the topological properties of the original graph.  
To summarize the main contributions of our paper:
\begin{enumerate}
    \item We propose the first graph super-resolution method based on direct edge representation learning.
    \item We demonstrate a computationally efficient approach capable of leveraging shallow GNN models for learning in the edge space.
    \item We provide strong empirical evidence that super-resolving brain graphs in higher-order topological space ensures superior topological consistency.
\end{enumerate}

\begin{figure}[t!]
\centering
\includegraphics[width=12cm]{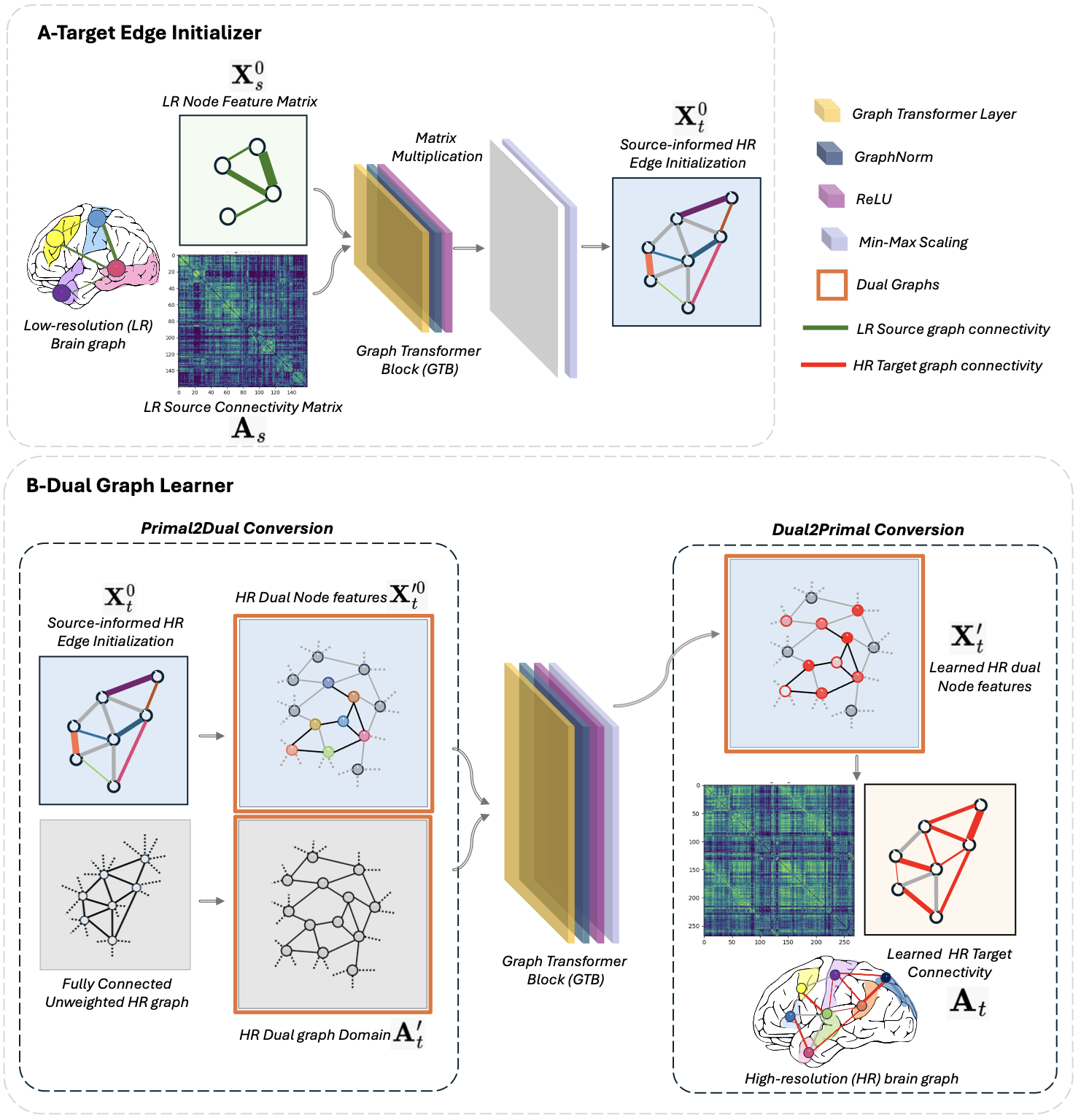}
\vspace{-10pt}
\caption{\footnotesize{ \emph{Illustration of the proposed \textbf{S}trongly \textbf{T}opology-\textbf{P}reserving GNN framework for Brain \textbf{G}raph \textbf{S}uper-\textbf{R}esolution (\textbf{STP-GSR}). \textbf{A-Target Edge Initializer.} We use the source graph to guide the initialization of our dual graphs. First, we pass each source graph $\mathcal{G}_s$ with adjacency matrix $\mathbf{A}_s$ and node feature matrix $\mathbf{X}^0_s$ through our GNN layer to learn rich node embeddings. Then, using matrix multiplication and min-max scaling, the node embeddings are transformed into a scalar edge feature matrix $\mathbf{X}^0_t$ for the target graph. In the next stage, these features are converted to dual node features. \textbf{B-Dual Graph Learner.} We use Primal2Dual conversion to map the learned edges to dual nodes, i.e., we flatten the upper triangular part of  $\mathbf{X}^0_t$ to get the dual node feature matrix $\mathbf{X}'^0_t$. We then perform node feature learning using our GNN layer in the dual space. Finally, we apply Dual2Primal conversion by mapping the learned dual node features to the upper triangular part of an empty target adjacency matrix, followed by reflection along the diagonal. This gives us the symmetric $\mathbf{A}_t$, which is our predicted connectivity matrix.} }}
\label{fig:1}
\end{figure}

\section{Methodology}
In this section, we describe our \textbf{S}trongly \textbf{T}opology-\textbf{P}reserving GNN framework for Brain \textbf{G}raph \textbf{S}uper-\textbf{R}esolution (\textbf{STP-GSR}) (\textbf{Figure \ref{fig:1}}). The key component of our model is the dual-graph formulation, which maps the edges of our brain graphs to nodes in a dual graph. This transformation simplifies the more challenging edge regression problem into an easier node regression task. The STP-GSR framework operates in two main stages. In the first stage, the \textit{TargetEdgeInitializer} encodes information from the LR source graph to initialize edge features of our HR target graph. In the second stage, the \textit{DualGraphLearner} constructs a dual graph and maps the HR edge features to its node space.  It then applies a GNN to update these features, followed by an inverse mapping from node space back to edge space, ultimately predicting the HR target connectivity matrix.

\begin{table}[t!]
    \centering
    \begin{tabular}{c|c|c}
    \hline
    Notation & Dimension & Definition\\
    \hline
    $n_s$& $\mathbb{N}$ & Number of nodes/ ROIs in the LR brain graph\\
     $n_t$   & $\mathbb{N}$ & Number of nodes/ ROIs in the HR brain graph\\
    $n'_t$   & $\mathbb{N}$ & Number of nodes in the dual graph\\
     $N$ & $\mathbb{N}$ & Number of samples in the training dataset\\
     $\mathbf{V}_s$ & - & The set of nodes/ ROIs in the LR brain graph\\
     $\mathbf{V}_t$ & - & The set of nodes/ ROIs in the HR brain graph\\
     $\mathbf{E}_s$ & - & The set of edge values/ connection strengths in the LR brain graph\\
     $\mathbf{E}_t$ & - & The set of edge values/ connection strengths in the HR brain graph\\
    $\mathbf{A}_s$ & $\mathbb{R}^{n_s \times n_s}$& The adjacency/ connectivity matrix for the LR brain graph\\
    $\mathbf{A}_t$ & $\mathbb{R}^{n_t \times n_t}$ & The adjacency/ connectivity matrix for the HR brain graph\\
    $\mathbf{A}'_t$ & $\mathbb{R}^{n'_t \times n'_t}$ & The adjacency matrix for the dual graph\\
     $\mathbf{X}_s^0$ & $\mathbb{R}^{n_s \times n_s}$& The initial node feature matrix for the LR brain graph\\
     $\mathbf{X}_s^1$ & $\mathbb{R}^{n_s \times n_t}$& The updated node feature matrix for the LR brain graph\\
    $\mathbf{X}_t^0$ & $\mathbb{R}^{n_t \times n_t}$& The learned initial edge feature matrix for the HR brain graph\\
    $\mathbf{X}'^0_t$ & $\mathbb{R}^{n'_t \times 1}$ & The initial node feature matrix for the dual graph\\
    $\mathbf{X}'_t$ & $\mathbb{R}^{n'_t \times 1}$ & The updated node feature matrix for the dual graph\\
    \hline
    \end{tabular}
    \caption{Major mathematical notations used in the paper}
    \label{tab:mathematical-notation}
\end{table}

\textbf{Problem statement}. Let $\mathcal{G}(\textbf{V}, \textbf{E}, \mathbf{A})$ denote a brain graph, where the elements of set $\textbf{V}$ represent the nodes or regions of interest (ROI) in the brain, the elements of set $\textbf{E}$ represent the connectivity strength between any two ROIs, and $\textbf{A}$ represents the connectivity matrix for the brain graph. Our training dataset consists of $N$ source-target pairs $(\mathcal{G}_s(\mathbf{V}_s, \mathbf{E}_s, \mathbf{A}_s), \mathcal{G}_t(\mathbf{V}_t, \mathbf{E}_t, \mathbf{A}_t))$ where $\mathcal{G}_s$ is the low-resolution (LR) brain graph, while $\mathcal{G}_t$ is the high-resolution (HR) brain graph. Additionally, each LR graph contains $n_s$ nodes (ROIs), while each HR graph consists of $n_t$ nodes with $n_s < n_t$. Our objective is to predict $\mathbf{A}_t$ given $\mathbf{A}_s$. 

\textbf{Primal-Dual Formulation}: We now introduce the key contribution of our paper, which enables direct computations and learning in the higher-order edge-space.

\emph{\textbf{Definition1}}. Given a directed graph $\mathcal{G}(\mathbf{V}, \mathbf{E}, \mathbf{A})$, also known as the primal graph, its dual graph $\mathcal{G}'(\mathbf{V}', \mathbf{E}', \mathbf{A}')$ is constructed as follows:

\begin{enumerate}
    \item Each node of the dual graph $\mathcal{G}'$ corresponds to an edge of the primal graph $\mathcal{G}$ i.e. $(i, j) \in \textbf{V}' \implies (i, j) \in \textbf{E}$ 
    \item Two dual nodes $(i, j), (k, l) \in \mathbf{V}'$ are connected by an edge in $\mathcal{G}'$ if they share a common direction and at least one common node in $\mathcal{G}$, i.e., the edges $(i, j)$ and $(k, l)$ have a common endpoint in $\mathcal{G}$
    \item Let $r$ and $c$ be the indices of the dual node $(i, j)$ and $(k, l)$, respectively. Then, $\textbf{A}'$ is an unweighted adjacency matrix s.t. $A'_{rc}=1$ if $(i, j)$ and $(k, l)$ are connected; otherwise $A'_{rc}=0$
\end{enumerate}

The dual graph defined above is also known as a line (di)graph or adjoint graph in graph theory \cite{gross2018graph}. For simple graphs, this primal-to-dual conversion is invertible, allowing us to easily map the dual nodes back to primal edges. 

Since our brain graphs are undirected, their connectivity matrices are symmetric, and it suffices to predict only the upper triangular part of $\mathbf{A}$. Therefore, we define a reduced edge-set $\mathbf{\hat{E}} = \{(i, j) | (i, j) \in \mathbf{E};i< j\}$ and simplify our dual graph formulation to:

\emph{\textbf{Definition 2}}. Given an undirected graph $\mathcal{G(\mathbf{V}, \mathbf{\hat{E}}, \mathbf{A})}$, it's dual graph $\mathcal{G}'(\mathbf{V}', \mathbf{E}', \mathbf{A}')$ is constructed as:
\begin{enumerate}
    \item Each node of the dual graph $\mathcal{G}'$ corresponds to an edge of the primal graph $\mathcal{G}$ i.e. $(i, j) \in \textbf{V}' \implies (i, j) \in \mathbf{\hat{E}}$ 
    \item Two dual nodes $(i, j), (k, l) \in \mathbf{V}'$ are connected by an edge in $\mathcal{G}'$ if they share a common node in $\mathcal{G}$ i.e. they satisfy the condition $i=k \cup i=l \cup j=k \cup j=l$
    \item Let $r$ and $c$ be the indices of the dual node $(i, j)$ and $(k, l)$ respectively. Then, $\textbf{A}'$ is an unweighted adjacency matrix s.t. $A'_{rc}=1$ if $(i, j)$ and $(k, l)$ are connected; otherwise $A'_{rc}=0$
\end{enumerate}

\emph{\textbf{Computational Efficiency}}. The above reformulation allows us to significantly reduce the computational requirements for the dual graph. For a simple primal graph with $n$ nodes, the dual graph from \textit{Definition 1} will have $n'= n(n-1)$ nodes in the worst case. This reduces by half for our simple undirected graphs in \textit{Definition 2}. Moreover, even though the number of dual nodes increases quadratically, the resulting $\textbf{A}'$ is highly sparse ( more than 97\%) even for our fully-connected primal graphs. This allows us to leverage the in-built optimizations in deep learning packages for highly sparse matrices.

\emph{\textbf{Topology-Preservation}}. The dual graph formulation offers two key topological advantages. First, existing GNN layers perform node representation learning and indirectly predict edge information, e.g., by taking the dot product of the node representations. This necessitates using computationally expensive GNN architectures to learn complex node representations capable of predicting edge features, which are not scalable to larger graphs. However, by shifting the computation to the edge-space, we learn directly on the edges, allowing the use of simpler GNN models. Second, nodes are considered 0-dimensional topological objects, while edges are 1-dimensional topological objects. Therefore, computing on these higher-order objects better captures higher-order relationships that underlie a diverse set of topological properties.

\textbf{Graph Transformer Block (GTB)}. This block encapsulates the key graph computational elements and is composed of:

\textit{Graph Transformer Layer}. We adopt the implementation in \cite{shi2020masked} for the graph transformer layer. For an input graph $\mathcal{G}_s(\mathbf{V}_s, \mathbf{E}_s, \mathbf{A}_s)$, let the input node feature matrix be $\mathbf{X}^0_s=\mathbf{A}_s$, i.e., the connectivity pattern for each node becomes its initial feature vector ($\mathbf{X}^0_s=[\mathbf{x}_1, \mathbf{x}_2, ..., \mathbf{x}_n]^T$). Next, for each node $i$, its representation is updated as:




\begin{align}
    \hat{\mathbf{x}}_i &= \mathbf{W}_0[\hat{\mathbf{x}}_i^1 || \hat{\mathbf{x}}_i^2|| ...|| \hat{\mathbf{x}}_i^H]
\end{align}

\begin{align}
    \hat{\mathbf{x}}_{i}^{h} &= \textbf{W}_1^h\textbf{x}_i + \sum_{j \in \mathcal{N}(i)} \alpha_{ij}^h (\mathbf{W}_2^h\mathbf{x}_j + \mathbf{W}_6^h\mathbf{A}_{ij})
\end{align}

\begin{align}
    \alpha_{ij}^h &= softmax\Bigg( \dfrac{(\textbf{W}_3^h\textbf{x}_i)^T (\textbf{W}_4^h\textbf{x}_j + \textbf{W}_6^h)\textbf{A}_{ij}}{\sqrt{d}}\Bigg)
\end{align}

where, $\hat{\mathbf{x}}_{i}^{h}$ is the output from head $h$, $\alpha_{ij}^h$ is the attention coefficient, $\{\mathbf{W}_k^h | k \in \{1, 2, ...,6\}\} \cup \{\mathbf{W}_0\}$ are learnable parameters, and $d$ is the dimension of $\mathbf{x}_i$.

We chose the graph transformer layer because it allows us to dynamically weight the importance of neighboring nodes. It uses the highly efficient multi-head self-attention mechanism to generate these dynamic weights and aggregate information. The learned representations are more expressive than GCNConv \cite{kipf2016semi} while being more computationally efficient than NNConv \cite{simonovsky2017dynamic}, two widely used layers in graph super-resolution.

\textit{GraphNorm}. After calculating the updated representations for graph $\mathcal{G}_s$, we apply instance based normalization \cite{cai2021graphnorm}, which helps stabilize and accelerate training. 

\textit{Non-linearity}. Earlier methods used sigmoid to enforce that the predicted connectivity strengths $\in [0, 1]$. However, we observed that this over-saturates the output and leads to vanishing gradients. Therefore, we switched to ReLU.


\textbf{A-TargetEdgeInitializer}. This block distills information from the source LR primal graph $\mathcal{G}_s$ to generate initial edge features for the target HR graph. Specifically, given $\mathbf{A}_s$ and $\mathbf{X}_s$ for a source LR graph, we update the LR node features as $\mathbf{X}_s^1=GTB(\mathbf{A}_s, \mathbf{X}_s)$, followed by  matrix multiplication to generate scalar edge features  as $\mathbf{X}_t^0={\mathbf{X}_s^1}^T \mathbf{X}_s^1$.

\textbf{B-DualGraphLearner}. This block uplifts the GNN operations to edge space and predicts the final target HR connectivity $\textbf{A}_t$. First, it creates a HR dual graph ${\textbf{A}'}_t$ and maps the learned $\mathbf{X}_t^0$ to the dual space using the  \textit{Primal2Dual} conversion ($\textbf{X}_t^0 \mapsto {\textbf{X}'}_t^0$). Then, it refines the edge features in the dual space as ${\mathbf{X}'}_t = GTB({\textbf{X}'}_t^0, {\textbf{A}'}_t)$. Finally, it applies the \textit{Dual2Primal} conversion to transform $\mathbf{X}'_t$ back to the HR primal space and obtain the predicted target adjacency matrix $\mathbf{A}_t$.

\textit{Primal2Dual}. To map the edge features of our HR brain graphs to the node features of the dual graph,  we extract the upper triangular section of $\mathbf{X}_t^0$ and reshape it to get the one-dimensional dual node feature matrix ${\mathbf{X}'}_t^0$. However, we still need to define the underlying domain ${\mathbf{A}'}_t$ of the target dual graph. For this, we use a simple fully-connected unweighted graph with $n_t$ nodes and create its dual using the formulation in \textit{Definition 2}. ${\mathbf{A}'}_t$ defined this way assumes a maximally connected HR domain, while the specific edge features or connectivity strengths are learned by updating ${\mathbf{X}'}_t^0$.

\textit{Dual2Primal}. To map dual node features back to the edges of our target HR graph, we treat the values in $\textbf{X}'_t$ as the values in the upper triangular part of $\textbf{A}_t$. To get our final predicted target HR matrix $\textbf{A}_t$, we simply pad, reshape, and reflect $\textbf{X}'_t$ into a square matrix of size $n_t$. It is worth noting that even though we used our dual formulation to learn scalar edge values, it is capable of learning multi-dimensional edge features and could predict multiple connectivity properties at once. Finally, the whole framework was trained end-to-end using L1 loss between predicted $\mathbf{A}_t$ and the true target $\mathbf{A}_t$.

\section{Experimental Results and Discussion}

%
\textbf{Evaluation Dataset}. We evaluate our model on 167 subjects from the publicly available Southwest University Longitudinal Imaging Multimodal (SLIM) dataset \cite{liu2017longitudinal}. SLIM provides a set of structural, diffusion, and resting-state fMRI images along with rich behavioral data. We use the rfMRI scans to generate our LR and HR brain graphs. First, we preprocess the scans using the Preprocessed Connectomes Project Quality Assessment Protocol. We then parcellate them into different regions of interests (ROIs) using the Dosenbach \cite{dosenbach2010prediction} and Shen\cite{shen2013groupwise} brain atlases to obtain brain graphs of resolution $160 \times 160$ (LR) and $268 \times 268$ (HR), respectively. 

\textbf{Comparison methods}. We use a combination of existing and newly created baselines to benchmark our framework including:

\emph{\textbf{Adapted IMANGraphnet}}. IMANGraphNet \cite{mhiri2021non} is the current state-of-the-art GNN model on brain graph super-resolution. It super-resolves a $35 \times 35$ morphological connectivity matrix into $160 \times 160$ Dosenbach-parcellated rfMRI brain graph matrix. However, it uses the computationally expensive NNConv \cite{simonovsky2017dynamic} layers, which cannot be scaled to our larger dataset due to out-of-memory (OOM) errors. Therefore, we evaluate against a modified version of IMANGraphNet, which linearly projects the LR node feature matrix $\mathbf{X}_s$ to a lower-dimensional space before applying the NNConv layers. To maintain dimensional consistency, we apply another linear projection to map the output back to a higher-dimensional space.

\emph{\textbf{Direct SR}}. To isolate the impact of our Primal-Dual formulation, we define a \textit{DirectSR} model that uses the Graph Transformer Block (GTB) to directly super-resolve the LR matrices into HR matrices. To ensure a fair comparison, we also increase its capacity by adding an additional graph transformer layer. 

\emph{\textbf{LR-HR-LR AutoEncoder}}. Inspired by the iterative up-and-down sampling methods in image super-resolution \cite{haris2018deep}, we propose an autoencoder model to capture the mutual dependency of LR and HR graphs. This model uses a GTB block to super-resolve a LR matrix to HR followed by another GTB block to down-resolve this HR matrix to LR. The model is trained using both the prediction loss for the HR matrix and the reconstruction loss for the LR matrix. 

\textbf{Parameter Setting}. For all the models, we use a batch size of 16 along with the gradient accumulation trick to avoid memory issues. For all the models except \textit{Adapted IMANGraphNet}, we use a learning rate of 0.005. For \textit{Adapted IMANGraphNet}, following the original paper \cite{mhiri2021non}, we use a learning rate of 0.025. We also apply binary search to select the largest low-dimensional space ($dim=32$) that does not cause OOM errors. For the graph transformer layer in the node space, we use 4 attentional heads with a dropout probability of 0.2. For the graph transformer layer in the edge-space, we use a single head as STP-GSR only uses a single graph transformer layer to predict scalar connectivity strengths. Finally, all models are trained for 60 epochs.

\textbf{Evaluation measures}. We perform three-fold cross-validation and report the average performance across all folds. In addition to the mean absolute error (MAE) between the true and predicted HR matrix, we evaluate our model using seven different topological measures. These measures were selected to assess a diverse set of brain connectivity patterns.

\begin{figure}[t!]
    \centering
    \includegraphics[width=12cm]{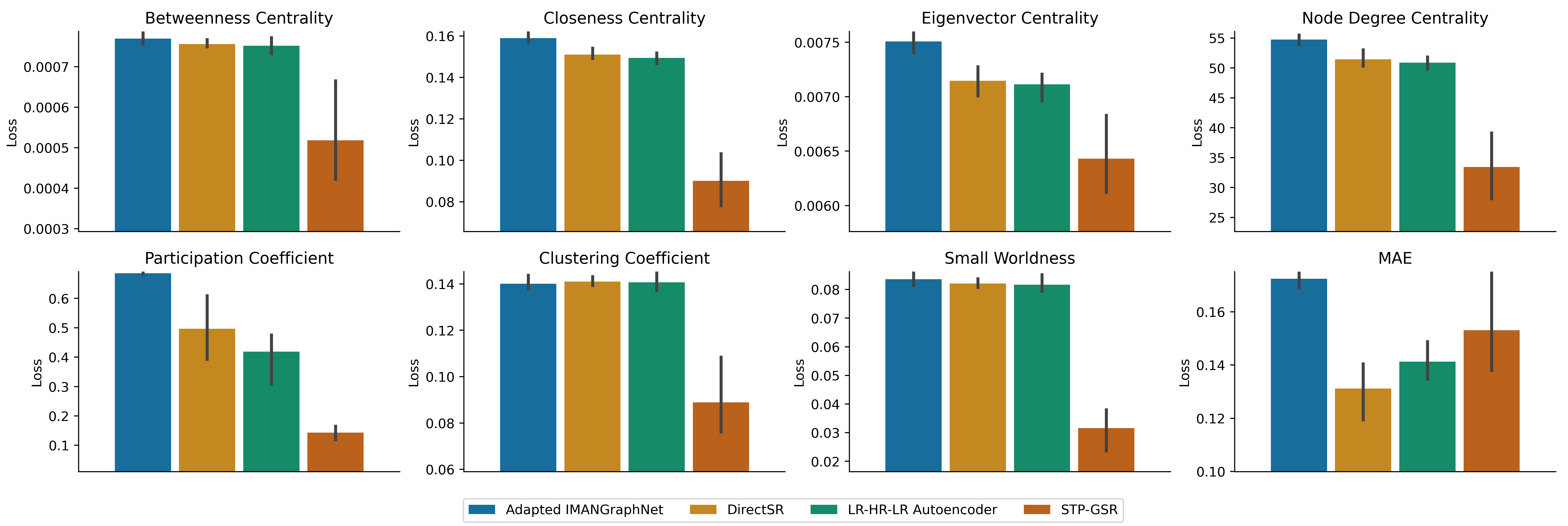}
    \caption{\emph{Performance results. We evaluate STP-GSR across 8 different metrics against two newly created baselines and IMANGraphNet. All losses are calculated w.r.t. ground truth HR graphs. We observe that STP-GSR consistently outperforms other methods on all topological metrics. However, it does struggle to perform on the mean absolute error (MAE). We suspect this to be due to the use of a very small and shallow GNN model on the edge space.}}
    \label{fig:2}
\end{figure}

\emph{\textbf{Relevance of topological measures}}. \textit{Node degree centrality} measures the number of incident connections to a given node and serves as an indirect measure of network resilience \cite{achard2006resilient}. \textit{Betweenness centrality} measures the fraction of shortest paths between all node pairs that pass through a given node and is useful for detecting bridge nodes between disparate regions \cite{rubinov2010complex}. \textit{Closeness centrality} quantifies the mean distance between a given node and the rest of the network, indicating the speed of communication within the network. \textit{Eigenvector centrality} assess the number of connections to a given node, weighted by the centrality of its neighbors, and evaluates hierarchical influence \cite{lorenzini2023eigenvector}. \textit{Participation Coefficient} and \textit{Clustering Coefficient} measures modularity in the network. The \textit{Participation Coefficient} measures the diversity of intermodular interconnections of individual nodes, while the \textit{Clustering Coefficient} assesses the presence of cliques or clusters. These metrics are important for evaluating brain network segregation and information processing within specialized brain subsystems \cite{gamboa2014working}. Finally, \textit{Small-worldness} is defined by the ratio between the characteristic path length and mean clustering coefficient (normalized by the corresponding values calculated on random graphs). It supports both segregated/specialized and distributed/integrated information processing \cite{watts1998collective}.

\begin{table}[]
    \centering
    \begin{tabular}{c|c}
    \hline
      Model   &  Number of model parameters ($10^{6}$) \\
      \hline
        Adapted IMANGraphNet & 0.099\\
        DirectSR & 1.103\\
        LR-HR-LR AutoEncoder & 2.205\\
        STP-GSR & 0.174\\
        \hline
    \end{tabular}
    \caption{Comparison of model size. Even though IMANGraphNet requires lesser number of parameters, its runtime memory usage is prohibitive.}
    \label{tab:params}
\end{table}

\emph{\textbf{Performance Analysis}}. From \textbf{Figure \ref{fig:2}}, we observe that STP-GSR consistently outperforms other methods on all topological measures. This demonstrates the advantage of our primal-dual formulation. Computations in the dual space benefit from propagating and aggregating information through higher-order topological objects (in this case, the edges) and inherently capture a diverse set of topological properties. While mapping to the dual space preserves topology,  the absolute error on the predicted HR connectivity strengths is determined by the richness of our \textit{TargetEdgeInitializer} and subsequent GNN computations in the HR dual space. However, as we are using a very small and shallow GNN on dual graphs, we observe that our model underperforms on the mean absolute error (MAE) metric. 

\section{Conclusion}
 
In this work, we introduced STP-GSR, the first graph super-resolution framework based on direct edge representation learning. Our main contributions include: (1)  A computationally efficient approach leveraging graph duality to map edge space computations to node space computations; (2) An end-to-end trainable pipeline integrating duality with brain graph super-resolution. We evaluated our framework against existing and newly created baselines across seven neurobiologically relevant topological measures, where it significantly outperformed the previous state-of-the-art model and newly created baselines.  Even though our model performs exceptionally well in preserving brain graph topology, its capacity to accurately predict the connectivity strength is limited by richness of the learned dual node features. As a future work, we aim to better understand the impact of dual node feature initialization and the possibility to synergetically combine node and edge space computations. 

\section{Supplementary Material}
We provide below supplementary material for reproducible and open science:
\begin{enumerate}
    \item A 5-mn YouTube video explaining how STP-GSR works on BASIRA YouTube channel at \url{https://www.youtube.com/watch?v=s1k-TcjKGFI}
    \item Python code on Github at \url{https://github.com/basiralab/STP-GSR}
\end{enumerate}

\bibliography{refs}
\bibliographystyle{splncs}
\end{document}